\documentclass{article}
\usepackage{booktabs}        
\usepackage{siunitx}       
\usepackage{amsmath}         
\usepackage{adjustbox}       
\usepackage{subcaption}      
\usepackage[normalem]{ulem} 
\usepackage{tikz}
\usepackage{pgfplots}
\pgfplotsset{compat=1.18}
\usepackage{subcaption}
\usepackage{adjustbox}
\usepackage{trimclip}
\usepackage{graphicx}
\usepackage{import}

\newcommand{\myparagraph}[1]{\noindent\textbf{#1}}
\newcommand{\ulnum}[1]{#1}

\usepackage{caption} 
\usepackage[final]{eiml_style_2025}

\title{Evaluating the Impact of Post-Training Quantization on Reliable VQA with Multimodal LLMs}
\author{%
  Paul Jonas Kurz$^{1*}$ \quad Tobias Jan Wieczorek$^{1}$ \quad Mohamed A. Abdelsalam$^{1}$\\
  \textbf{Rahaf Aljundi}$^{2}$ \quad \textbf{Marcus Rohrbach}$^{1}$\\
  $^1$TU Darmstadt \quad $^2$Toyota Motor Europe\\
}
\date{October 2025}

\begin{document}

\maketitle

\begin{abstract}
Multimodal Large Language Models (MLLM) are increasingly deployed in domains where both reliability and efficiency are critical. However, current models remain overconfident, producing highly certain but incorrect answers. At the same time, their large size limits deployment on edge devices, necessitating compression. We study the intersection of these two challenges by analyzing how Post-Training Quantization (PTQ) compression affects both accuracy and reliability in Visual Question Answering (VQA). We evaluate two MLLMs, Qwen2-VL-7B and Idefics3-8B, quantized with data-free (HQQ) and data-aware (MBQ) methods across multiple bit widths. To counteract the reduction in reliability caused by quantization, we adapt the Selector confidence estimator for quantized multimodal settings and test its robustness across various quantization levels and out-of-distribution (OOD) scenarios. We find that PTQ degrades both accuracy and reliability. Data-aware methods soften the effect thereof. The Selector substantially mitigates the reliability impact. The combination of int4$_{\textrm{MBQ}}$ and the Selector achieves the best efficiency-reliability trade-off, closing in on uncompressed performance at approx. \qty{75}{\percent} less memory demand. Overall, we present the first systematic study linking quantization and reliability in multimodal settings.
\end{abstract}

\vspace{-0.3em}
\section{Introduction}
Large multimodal models combining vision and language understanding now power both general-purpose assistants and specialized applications in healthcare, business, and accessibility. As these models are increasingly trusted with autonomous decision-making, their overconfidence becomes a major reliability concern~\cite{whitehead2022}. Reliable systems should \emph{abstain} when uncertain, following the principle of \textbf{selective prediction}~\cite{elYaniv2010}.

At the same time, there is a strong motivation to deploy MLLMs efficiently on resource-constrained devices, such as mobile or edge platforms. \textbf{Post-Training Quantization (PTQ)} reduces the memory and computation of trained models by mapping parameters to lower-precision formats~\cite{gholami2021}. However, it may distort internal representations and thus alter model confidence behavior.

While both reliability estimation and model compression are active areas of research, their intersection remains largely unexplored. Prior PTQ work focuses on task accuracy for unimodal LLMs~\cite{jin2024, li2024}, whereas selective prediction research assumes uncompressed networks~\cite{whitehead2022, dancette2023}. Understanding how quantization affects the reliability of multimodal perception and reasoning is essential for deploying trustworthy and efficient models.

We address this gap through a systematic empirical investigation of quantized MLLMs for selective VQA, using both intrinsic and Selector-based confidence estimation (see Sec.~\ref{sec:selective_prediction}). Our study is organized around three research questions:
\begin{enumerate}
\item \textbf{RQ1:} How does PTQ affect multimodal performance and reliability?  
\item \textbf{RQ2:} Can the Selector mitigate reliability loss introduced by quantization?  
\item \textbf{RQ3:} How robust is the Selector across quantization intensities and OOD conditions?
\end{enumerate}

We contribute the first multimodal evaluation of quantization effects in terms of both accuracy and reliability, comparing data-free and data-aware methods across a range of bit widths. We conduct an extensive study across the VQAv2~\cite{goyal2017}, AdVQA~\cite{sheng2021}, and VizWiz~\cite{gurari2018} datasets. We also present an analysis showing that losses in accuracy and reliability are correlated, with the Selector effectively compensating for drops in reliability. 

\vspace{-0.3em}
\section{Background and Related Work}

\textbf{Multimodal Large Language Models.}
Transformer architectures~\cite{vaswani2017} enabled unified vision–language reasoning via models such as BLIP-2~\cite{li2023}, LLaVA~\cite{hliu2023}, and Qwen2-VL~\cite{wang2024}. These MLLMs integrate frozen vision encoders with pretrained language decoders, enabling zero-shot multimodal understanding. VQA \cite{antol2015, goyal2017, marino2019, singh2019} remains the canonical benchmark for such systems.

\textbf{Selective Prediction.}
 A selective model either accepts the predictor's response or abstains based on the output of a confidence estimator \cite{elYaniv2010}. Intrinsic estimators (e.g., \textbf{MaxProb}~\cite{hendrycks2017}) are not explicitly separated from the predictive model, whereas extrinsic models (e.g., \textbf{Selector}~\cite{whitehead2022, dancette2023}) are. Selector-based selective VQA has been shown to improve calibration and reduce overconfident errors.

\textbf{Quantization.}
Post-Training Quantization (PTQ) compresses trained networks by mapping weights and activations to lower bit width data types \cite{gholami2021}. \textbf{Data-free PTQ} (e.g., Half-Quadratic Quantization, HQQ~\cite{badri2023}) optimizes scaling using heuristics, while \textbf{data-aware PTQ} (e.g., Modality-Balanced Quantization, MBQ~\cite{li2025}) uses calibration datasets to capture activation dynamics and determine optimal transformations. Both approaches can drastically reduce memory, but their effect on reliability in MLLMs remains unknown, previously only being touched on in unimodal settings \cite{jin2024, li2024}.

\textbf{Research Gap.}
No prior work systematically quantifies the interplay between quantization level and reliability in multimodal LLMs. This study provides the first empirical bridge between the two.

\vspace{-0.3em}
\section{Methodology}

\begin{table}[b!]
\centering
\scriptsize
\setlength{\tabcolsep}{3pt}
\caption{
Comparison of data-free (HQQ) and data-aware (MBQ) PTQ on VQAv2 for \textbf{Idefics3} and \textbf{Qwen2-VL}. 
We evaluate accuracy, calibration (ECE), and selective prediction ($\mathcal{C}@\mathcal{R}$, AUC, $\Phi_c$) with \textbf{MaxProb} confidence estimates. 
Best results are in \textbf{bold}.
}
\label{tab:max-prob-ptq-vqav2}
\begin{adjustbox}{max width=\linewidth}
\begin{tabular}{l
                S S S S S S S S
                S S S S S S S S}
\toprule
& \multicolumn{8}{c}{\textbf{Idefics3}} & \multicolumn{8}{c}{\textbf{Qwen2-VL}} \\
\cmidrule(lr){2-9} \cmidrule(lr){10-17}
&  {($\uparrow$)} &  {($\downarrow$)} 
& \multicolumn{3}{S}{{$\mathcal{C}@\mathcal{R}$ ($\uparrow$)}}  
& {($\downarrow$)} 
& \multicolumn{2}{S}{{$\Phi$ ($\uparrow$)}} 
&  {($\uparrow$)} &  {($\downarrow$)} 
& \multicolumn{3}{S}{{$\mathcal{C}@\mathcal{R}$ ($\uparrow$)}}  
& {($\downarrow$)} 
& \multicolumn{2}{S}{{$\Phi$ ($\uparrow$)}} 
\\
\cmidrule(lr){2-2}\cmidrule(lr){3-3} \cmidrule(lr){4-6}\cmidrule(lr){7-7} \cmidrule(lr){8-9}
\cmidrule(lr){10-10}\cmidrule(lr){11-11} \cmidrule(lr){12-14}\cmidrule(lr){15-15} \cmidrule(lr){16-17}
Quant. ($f$)
& {Acc.} & {ECE} & {\qty{0.5}{\percent} } & {\qty{1}{\percent}} & {\qty{5}{\percent} } & {AUC } & {$\Phi_{10}$ } & {$\Phi_{100}$}
& {Acc.} & {ECE} & {\qty{0.5}{\percent} } & {\qty{1}{\percent}} & {\qty{5}{\percent} } & {AUC} & {$\Phi_{10}$} & {$\Phi_{100}$ } \\
\midrule
bf16
& \bfseries 79.3 & \ulnum{5.1} & \bfseries 14.2 & 23.6 & \ulnum{53.3} & \bfseries 6.3 & \bfseries 37.2 & \ulnum{12.3}
& \bfseries 83.0 & \bfseries 3.5 & \bfseries 31.8 & \ulnum{42.6} & \bfseries 70.4 & \bfseries 3.8 & \ulnum{51.1} & \bfseries 25.8 \\
\midrule
int8$_\text{HQQ}$
& \bfseries 79.3 & \ulnum{5.1} & 13.8 & \bfseries 24.0 & \bfseries 53.7 & \bfseries 6.3 & \ulnum{37.1} & 12.3
& \ulnum{82.9} & \bfseries 3.5 & \bfseries 31.8 & \bfseries 42.7 & \bfseries 70.4 & \bfseries 3.8 & 51.0 & \bfseries 25.8 \\

int8$_\text{MBQ}$
& \bfseries 79.3 & \ulnum{5.1} & \ulnum{14.0} & \ulnum{23.9} & \ulnum{53.3} & \bfseries 6.3 & 37.0 & \bfseries 12.4
& \bfseries 83.0 & \bfseries 3.5 & \bfseries 31.8 & \ulnum{42.6} & \bfseries 70.4 & \bfseries 3.8 & \bfseries 51.3 & \ulnum{25.5} \\

int4$_\text{HQQ}$
& 77.2 & 6.1 & 7.2 & 16.3 & 48.3 & 7.4 & 32.6 & 8.4
& 82.2 & 4.0 & \ulnum{28.3} & 39.5 & 68.1 & 4.09 & 48.9 & 23.7 \\

int4$_\text{MBQ}$
& \ulnum{78.1} & 5.2 & 8.1 & 18.0 & 49.3 & \ulnum{7.0} & 33.7 & 9.3
& 82.5 & \ulnum{3.7} & 27.9 & 41.1 & \ulnum{69.2} & \ulnum{3.9} & 50.3 & 23.6 \\

int3$_\text{HQQ}$
& 64.0 & 7.9 & 0.0 & 0.1 & 8.0 & 17.4 & 7.5 & -0.3
& 80.1 & 9.1 & 9.3 & 21.2 & 58.2 & 5.5 & 41.0 & 10.9 \\

int3$_\text{MBQ}$
& 71.2 & \bfseries 3.9 & 0.6 & 7.2 & 33.5 & 10.6 & 22.5 & 3.3
& 81.2 & 4.6 & 28.0 & 37.7 & 65.2 & 4.5 & 46.5 & 21.3 \\
\bottomrule
\end{tabular}
\end{adjustbox}
\end{table}

\subsection{Selective VQA with Multimodal LLMs}
\label{sec:selective_prediction}
A selective VQA model $h=(f,g)$ answers the input visual question $x$ if and only if $g(x)\ge \gamma$, where $f(x)$ is the VQA model and $g(x)$ a confidence estimator. The optimal $g$ ranks samples by correctness probability, ensuring true loss monotonicity \cite{geifman2017}.

\paragraph{MaxProb (Intrinsic Baseline).}
For autoregressive decoders, intrinsic confidence can be defined as the joint softmax probability of all generated tokens \cite{dancette2023}:
\begin{equation}
g(x)=\prod_{t_k\in t[1:n]}p_\theta(t_k\mid x,t_{<k}).
\end{equation}
Higher joint probabilities usually indicate more reliable answers, but this measure is often overconfident and poorly calibrated \cite{uncalibrated_llm}.

\paragraph{Selector (Extrinsic Estimator).}
Following~\cite{dancette2023}, we train a two-layer MLP that predicts the likelihood of correctness from multimodal signals: the max-pooled representations of image ($v_i$) and question ($q_i$), the multimodal embedding used for generating the first output token ($o_1$), and the joint output token probability ($p$). Selector regresses on non-binary VQA accuracy targets during training, thus approximating the optimal selection function that yields effective abstention behavior.

\vspace{-0.3em}
\subsection{Quantization Methods}

We evaluate two PTQ schemes representing the trade-off between accuracy and efficiency.

\myparagraph{HQQ (Half-Quadratic Quantization)~\cite{badri2023}.}
A \emph{data-free} method designed for large models. It assumes quantization errors to follow a hyper-Laplacian distribution, optimizing scale and zero-point via a closed-form half-quadratic solver. Efficient but more susceptible to activation outliers.

\myparagraph{MBQ (Modality-Balanced Quantization)~\cite{li2025}.}
A \emph{data-aware} method tailored for MLLMs. It leverages channel-wise equalization weighted by gradient magnitudes of vision and text tokens, thereby applying modality-aware conditioning to quantization targets. Calibration uses \num{128} samples from the ShareGPT4V~\cite{chen2024} dataset. MBQ is more costly than HQQ but preserves multimodal sensitivity and accuracy across bit widths better.

Both methods are evaluated at int8, int4, and int3 precision in weight-only mode.

\vspace{-0.3em}
\subsection{Experimental Setup}

\myparagraph{Models.} We use Qwen2-VL-7B~\cite{wang2024} and Idefics3-8B~\cite{laurencon2024}, both loaded in bf16 precision and quantized post-hoc. Generation uses greedy decoding.

\myparagraph{Datasets.} We evaluate on VQAv2~\cite{goyal2017} (in-distribution), AdVQA~\cite{sheng2021} (linguistic OOD), and VizWiz~\cite{gurari2018} (multimodal OOD). The Selectors are trained on a set fraction of VQAv2 validation data, following prior work \cite{whitehead2022}, and evaluated on all datasets.

\myparagraph{Metrics.} We report accuracy, Expected Calibration Error (ECE), and selective prediction metrics: Coverage-at-Risk ($\mathcal{C}@\mathcal{R}$) at \qty{0.5}{\percent}, \qty{1}{\percent}, and \qty{5}{\percent} risk levels, Area under the Risk–Coverage Curve (AUC), and Effective Reliability ($\Phi_c$)~\cite{whitehead2022}. Thresholds for $\Phi_c$ are chosen on a held-out VQAv2 split.

\vspace{-0.3em}
\section{Results and Analysis}

\subsection{RQ1: Effect of Quantization}

\begin{table}[tb!]
\centering
\scriptsize
\setlength{\tabcolsep}{3pt}
\caption{
Comparison of \textbf{Selector} and \textbf{MaxProb} confidence estimates under data-free (HQQ) and data-aware (MBQ) PTQ on VQAv2 for \textbf{Idefics3} and \textbf{Qwen2-VL}.
We evaluate calibration (ECE) and selective prediction ($\mathcal{C}@\mathcal{R}$, AUC, $\Phi_c$). 
Best results are in \textbf{bold}.
}
\label{tab:selector-ptq-comparison}

\begin{adjustbox}{max width=\linewidth}
\begin{tabular}{l l
                S S S S S S S
                S S S S S S S}
\toprule
& & \multicolumn{7}{c}{\textbf{Idefics3}} & \multicolumn{7}{c}{\textbf{Qwen2-VL}} \\
\cmidrule(lr){3-9} \cmidrule(lr){10-16}
& &  {($\downarrow$)} 
& \multicolumn{3}{S}{{$\mathcal{C}@\mathcal{R}$ ($\uparrow$)}}  
& {($\downarrow$)} 
& \multicolumn{2}{S}{{$\Phi$ ($\uparrow$)}} 
&  {($\downarrow$)} 
& \multicolumn{3}{S}{{$\mathcal{C}@\mathcal{R}$ ($\uparrow$)}}  
& {($\downarrow$)} 
& \multicolumn{2}{S}{{$\Phi$ ($\uparrow$)}} 
\\
\cmidrule(lr){3-3}\cmidrule(lr){4-6}\cmidrule(lr){7-7} \cmidrule(lr){8-9}
\cmidrule(lr){10-10} \cmidrule(lr){11-13}\cmidrule(lr){14-14} \cmidrule(lr){15-16} 
Name & Quant. ($f$)
& {ECE} & {\qty{0.5}{\percent}} & {\qty{1}{\percent}} & {\qty{5}{\percent}} & {AUC} & {$\Phi_{10}$ } & {$\Phi_{100}$}
& {ECE} & {\qty{0.5}{\percent}} & {\qty{1}{\percent}} & {\qty{5}{\percent}} & {AUC} & {$\Phi_{10}$} & {$\Phi_{100}$} \\
\midrule
MaxProb & bf16
& 5.1 & 14.2 & 23.6 & 53.3 & 6.3 & 37.21 & 12.3
& 3.5 & 31.8 & 42.6 & 70.4 & 3.8 & 51.1 & 25.8 \\
\midrule
Selector & bf16
& \bfseries 1.4 & 27.6 & \ulnum{36.3} & \ulnum{62.2} & \bfseries 5.1 & \bfseries 44.3 & 20.4
& 3.0 & \ulnum{37.4} & \ulnum{46.4} & \bfseries 72.1 & \bfseries 3.6 & \bfseries 53.7 & \bfseries 30.9 \\
\midrule
Selector & int8$_\text{HQQ}$
& 3.4 & \bfseries 27.8 & 35.7 & 62.1 & \bfseries 5.1 & 44.1 & \ulnum{21.6}
& 4.3 & \bfseries 37.9 & \bfseries 46.5 & \ulnum{72.0} & \bfseries 3.6 & \ulnum{53.6} & \ulnum{30.5} \\

Selector & int8$_\text{MBQ}$
& \ulnum{1.9} & \ulnum{27.7} & \bfseries 36.4 & \bfseries 62.3 & \bfseries 5.1 & \bfseries 44.3 & \bfseries 22.4
& 3.2 & 37.3 & 45.9 & \bfseries 72.1 & \bfseries 3.6 & 52.9 & 30.0 \\

Selector & int4$_\text{HQQ}$
& 3.2 & 22.3 & 30.3 & 56.1 & 6.2 & 39.8 & 19.0
& \ulnum{2.2} & 35.0 & 43.7 & 69.9 & 3.9 & 51.5 & 28.4 \\

Selector & int4$_\text{MBQ}$
& 2.6 & 24.1 & 32.9 & 58.5 & \ulnum{5.7} & \ulnum{41.3} & 19.0
& \ulnum{2.2} & 35.4 & 45.4 & 71.0 & \ulnum{3.7} & 51.9 & 28.9 \\

Selector & int3$_\text{HQQ}$
& 7.0 & 3.1 & 7.0 & 26.0 & 14.4 & 17.2 & 3.4
& 4.3 & 27.9 & 37.1 & 63.9 & 4.8 & 45.7 & 22.3 \\

Selector & int3$_\text{MBQ}$
& \bfseries 1.4 & 13.2 & 19.7 & 44.0 & 9.1 & 29.8 & 11.9
& \bfseries 1.7 & 32.2 & 40.8 & 66.8 & 4.3 & 48.6 & 26.0 \\
\bottomrule
\end{tabular}
\end{adjustbox}
\end{table}

As shown in Table~\ref{tab:max-prob-ptq-vqav2}, quantization consistently reduces both task accuracy and reliability. As bit width decreases, accuracy drops and ECE increases, showing that model calibration deteriorates correspondingly. Data-aware MBQ maintains higher accuracy and lower calibration error than HQQ, especially at \num{4} bits and below. At int4, performance remains within about \num{2} percentage points of bf16, while int3 introduces severe confidence instability. Reliability degradation mirrors accuracy loss: quantization noise directly perturbs the confidence distribution.

\subsection{RQ2: Selector Compensation}
Selector significantly improves reliability for both quantized and bf16 models (see Table~\ref{tab:selector-ptq-comparison}). Compared to MaxProb, the Selector consistently lowers ECE and improves $\mathcal{C}@$\qty{1}{\percent} across all bit widths, demonstrating better calibration and selective prediction capabilities under quantization. The Selector restores reliability for both models to values comparable to the bf16 baseline, except for int3 quantization, where intrinsic noise limits compensation. The Selector thus acts as an efficient reliability-restoration mechanism without modifying base model weights.

\subsection{RQ3: Out-of-Distribution Selector Robustness}
\vspace{-3pt}
\begin{figure}[tb!]
    \centering
    \includegraphics[trim=3.83cm 19cm 4.09cm 2.685cm, clip, width=\textwidth]{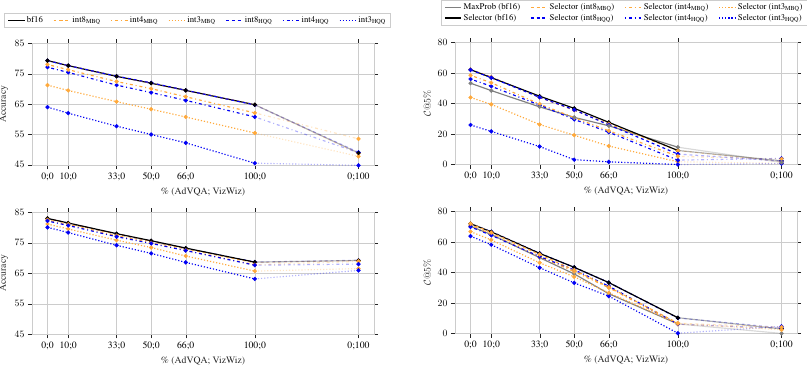}
    \caption{\textbf{Evolution of VQA accuracy and coverage across quantizations.}
    0;0 denotes full in-distribution data from VQAv2. OOD data is sourced from AdVQA and VizWiz. The opaque transition from 100;0 to 0;100 reflects the fundamental multimodal shift caused by a transition towards conversational questions and lower-quality images. Top is \textbf{Idefics3-8B}, bottom is \textbf{Qwen2-VL-7B}.}
    \label{fig:ood_Results}
\end{figure}
\vspace{-5pt}

As shown in Figure \ref{fig:ood_Results}, shifting from VQAv2 to AdVQA and VizWiz introduces progressively stronger multimodal distribution shifts that stress both quantized and bf16 models. Across datasets, performance and reliability deteriorate roughly in proportion to their in-distribution degradation, indicating that quantization amplifies but does not fundamentally alter the model’s robustness trends. Data-aware MBQ quantization consistently yields smoother declines and greater reliability retention than HQQ, especially under moderate shifts. The Selector enhances coverage and effective reliability throughout this progression, delaying but not preventing the collapse that occurs under severe OOD conditions. Notably, Selector behavior remains tightly correlated with intrinsic confidences, suggesting that its benefit stems from stabilizing distorted activation patterns rather than learning an entirely independent uncertainty signal.

\vspace{-0.3em}
\section{Discussion and Conclusion}
Quantization provides substantial efficiency gains, as moving from bf16 to int4 reduces memory by roughly \qty{75}{\percent}.
However, this introduces a predictable decline in both accuracy and reliability. Our results show that this degradation is proportional rather than catastrophic: data-aware MBQ maintains better calibration and robustness than data-free HQQ, and a lightweight Selector recovers much of the lost reliability without model retraining. Pairing an int4$_{\textrm{MBQ}}$ model with a Selector offers the best balance between efficiency and dependability, preserving \qty{98}{\percent} of bf16 accuracy while sustaining near-identical calibration and selective performance. Future work includes exploring quantization-aware reliability training, modeling intrinsic uncertainty under compression, and extending these approaches to broader multimodal reasoning tasks.

\section*{Acknowledgements}
This research was partially funded by an Alexander von Humboldt Professorship in Multimodal Reliable AI, sponsored by Germany’s Federal Ministry of Research, Technology and Space and by a LOEWE Spitzen-Professur (LOEWE/4a//519/05.00.002(0010)/93).
We gratefully acknowledge support from the hessian.AI Service Center (funded by the Federal Ministry of Research, Technology and Space, BMFTR, grant no. 16IS22091) and the hessian.AI Innovation Lab (funded by the Hessian Ministry for Digital Strategy and Innovation, grant no. S-DIW04/0013/003).

\bibliographystyle{plain}
\bibliography{main}
\end{document}